# Dense Multiscale Feature Fusion Pyramid Networks for Object Detection in UAV-Captured Images

Yingjie Liu[1]

[1]North University of China

**Abstract:** Although much significant progress has been made in the research field of object detection with deep learning, there still exists a challenging task for the objects with small size, which is notably pronounced in UAV-captured images. Addressing these issues, it is a critical need to explore the feature extraction methods that can extract more sufficient feature information of small objects. In this paper, we propose a novel method called Dense Multiscale Feature Fusion Pyramid Networks(DMFFPN), which is aimed at obtaining rich features as much as possible, improving the information propagation and reuse. Specifically, the dense connection is designed to fully utilize the representation from the different convolutional layers. Furthermore, cascade architecture is applied in the second stage to enhance the localization capability. Experiments on the drone-based datasets named VisDrone-DET suggest a competitive performance of our method.

## 1. Introduction

With the rapid development of aerial technology, especially for UAVs which have found a wide range of applications in the commercial field, including agricultural, aerial photography, fast delivery, environmental monitoring, etc [1]. Consequently, more and more attention has been paid to the research of general computer vision algorithms, such as object detection in aerial images. In previous years, many works are mainly focused on the sliding window search [2] and the handcrafted features[3], which normally require a lot of prior knowledge and formula derivation. In recent years, object detection based on deep learning algorithms has become the dominant technique, these methods, such as R-CNN series [4-6], YOLO series [7-10], SSD series [11,12], etc, have achieved great success in natural image detection(e.g., images in Pascal VOC [13], MS COCO [14]). However, these approaches have been to result in undesirable performance when detecting objects in images or videos captured from UAVs. [15]

UAVs (or Drones) have been usually deployed in the large scene, that means there are lots of objects in a single image and most of these objects are very small size, which is a big characteristic for aerial images and remains an open challenge. Generally, some detectors, such as Faster R-CNN, SSD, and YOLO, only utilize the feature map from a single layer of CNN networks, which has limited the representation capability of the feature information. Recent works focus on feature fusion for object detection, Feature Pyramid Network (FPN) [16] is one of a classical method that combines low-level and high-level features information by adopting a top-down architecture and lateral connections. This typical feature fusion method greatly improves the detecting performance for objects with small size, since the low-level features have sufficient location information, which is quite important to small objects for both classification and localization. However, considering that each pyramid layer of FPN only focuses on the lateral connections from the corresponding feature map as shown in the lower part of figure 1, feature information of these other layers is still available to utilize and it is necessary to extract more contextual semantic information for small objects.

To obtain more sufficient representations of feature maps, we propose a novel feature fusion method named Dense Multiscale Feature Fusion Pyramid Networks (DMFFPN) to more efficiently fuse the low-level and high-level features information for feature maps. Besides, cascade architecture is used to refine the bounding box prediction in the second stage. A detailed outline of our framework is presented in Section 3. In general, the main contributions of this work are as follows:

1. We design a simple but effective feature fusion method called DMFFPN, which can fully exploit feature propagation, feature reuse, and enhance the performance for the prediction of objects, especially for the small objects in aerial images.

2. We adopt Cascade architecture in the second stage to refine the bounding box regression and overcome the difficulty of locating small objects in complex backgrounds.

3. Extensive experiments on the aerial image datasets named VisDrone [17] demonstrates the validity and stability of the proposed framework.

The rest of this paper is organized as follows. Section 2 gives the related work of natural image detection and small object detection. Section 3 shows the analysis and description of the proposed framework. Section 4 presents experiments of the drone-based dataset. The last section concludes this paper.

## 2. Related Work

Nowadays, Object detection has made hot interests among researchers in the field of computer vision. Generally, Traditional detection methods usually need much prior knowledge and complex formula derivations and thus have defects of generalization ability. But CNN-based object detection methods can learn feature information automatically and are more likely to apply in real life because of the strong ability of generalization. In general, the detection methods based on deep learning can be generally divided into one-stage methods and two-stage methods. One-stage methods are good at high speed, including SSD, YOLO, and two-stage methods such as Fast-RCNN, Faster-RCNN, and R-FCN , have a good performance on accuracy.

RCNN [4] is the first successful work that leads the methods of deep learning into object detection, it adopts the selective search algorithm and uses SVM as the classifier. Fast R-CNN improves by using ROI pooling method. Based on that, Faster RCNN adopts RPN(region proposal networks) to generate proposals. However, Faster R-CNN makes the prediction only on the final feature map, which is unfavorable to detect the small objects.

Considering the drawback in Faster R-CNN, SSD utilizes the multi-level features to predict on multiscale objects. Specifically, it uses low-level feature maps to predict the small objects and use high-level feature maps to predict the large objects, yet it does its prediction on the intermediate feature maps without using the shallow feature maps, which is important to detect small objects. General speaking, it is not beneficial to predict the small object on low resolution, because the large stride of feature extractor has made the semantic information of small objects vanished, so it is quite hard to detect its features in high-level feature maps.

In order to solve this problem, FPN addressed the problem in SSD by fusing the low-level and high-level features, including the bottom-up pathway, top-down pathway, and lateral connection. As everyone knows that the high-resolution maps have strong location information and low-resolution maps have strong semantic information, both of which are of vital importance for object detection especially for detecting small objects. Similar to the feature fusion methods of FPN, relevant works like RetinaNet [19] and Mask RCNN [20] also adopt the same structures as the baseline networks for better detection results. Recent works on feature fusion method HRNet [21] maintains high-resolution feature maps, gradually fuse high-to-low resolution sub-networks in parallel architecture to obtain rich representations. PANet [22] makes additional bottom-up path augmentation to shortens the transmission path of the feature information.

Compared with natural images, aerial images have many unique characteristics, such as small and densely distributed objects, which takes a large proportion in a single image. Because small objects have few pixels, which makes an unfavorable condition to the representation of feature information. Therefore, there is a hard difficulty for both classification and localization when detecting small objects and the detection results are still far from satisfactory. Our method combines different feature maps by adopting a dense connection to fully exploit the features of each layer, which can bring more contextual information and keep the small extra cost of computation.

## 3. Proposed Method

In this section, we will detail each part of the proposed work, which mainly include DMFFPN for the first stage and Cascade architecture for the second stage. Figure 1 gives the overall framework. Specifically, DMFFPN efficiently utilizes the feature information of each layer to generate the feature map that has a more powerful semantic representation. Next, getting proposals from the region proposal networks(RPN) for the second stage. Then, high-quality regression and classification of proposals are processed by Cascade R-CNN [23]. Finally, we get the detection result.

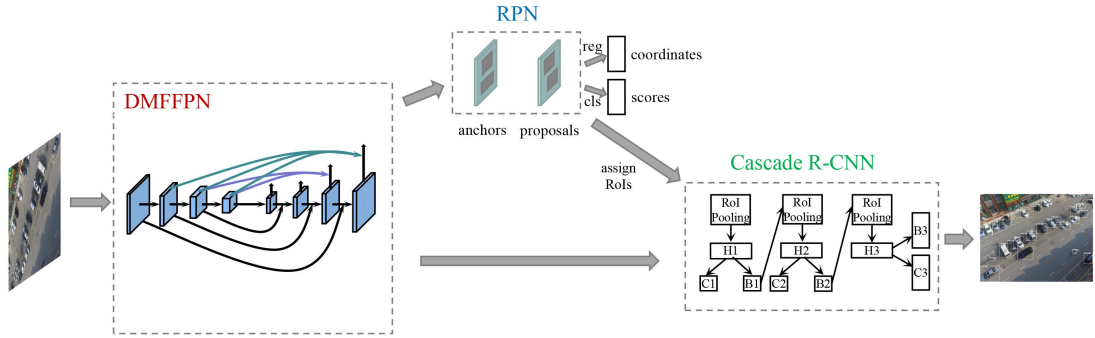

Figure 1. Illustration of overall network architecture.

### 3.1 Dense Multiscale Feature Fusion Pyramid Networks

As we all know, the low-level location information and high-level semantic information attach equal importance to object detection, especially to the aerial images detection, which has more small objects. FPN is an effective method that fusing multi-level information from low and high-level feature maps via the bottom-up, top-down, and lateral connection.

Since some objects like bicycle, tricycle and motor, both of them are generally smaller size compared with others. Besides, there are many similar classes such as car and van in aerial images, which can not be distinguished effectively. Moreover, the complexity of background increases the difficulty of the recognition process, for example, there are lots of vehicles-like disturbances due to the large scene of aerial images such as the shape of the roof, which is likely to confuse.

To solve the problems mentioned above, we propose a novel feature fusion method that makes a dense connection between Top-down pathway and bottom-up pathway. Figure 2 shows the architecture of DMFFPN based on ResNets [24], the details are as follows.

**Feature Pyramid Networks (FPN)**. We use the feature maps of the bottom-up pathway as $\{C_2, C_3, C_4, C_5\}$, which is from the last feature maps of each residual block with the strides $\{4, 8, 16, 32\}$ pixels. In top-down architecture, the pyramid layers are described as $\{P_2, P_3, P_4, P_5\}$.

**Dense Multiscale Feature Fusion**. The upper part described in figure 2 is the dense connection from the bottom-up pathway. We integrate all the valid and non-redundant connections to $\{P_2, P_3\}$ before finally predicting, note that we don't make connections to $\{P_4, P_5\}$, because there are none of the available connections from $\{C_2, C_3, C_4, C_5\}$. The specific definition is as follows.

$$P_5^* = Conv_{3\times3}[Conv_{1\times1}(C_5)] \tag{1}$$

$$P_4^* = Conv_{3\times3}[Conv_{1\times1}(C_4) + Upsample(P_5)] \tag{2}$$

$$P_i^* = Conv_{3\times3}\left\{\sum_{j=i+1}^{5} Upsample\left[Conv_{1\times1}(C_j)\right] \oplus P_i\right\} \tag{3}$$

where $Conv_{k\times k}$ represents the operation of convolution, which adopts kernel size 1×1 or 3×3. *Upsample* represents the up-sampling operation, in this paper, we apply bilinear up-sampling in all experiments. $\oplus$ is the operation of concatenation. $P_i^*$ is the final prediction of all fused feature maps from $C_j$ and $P_i$. So we can finally get $\{P_2^*, P_3^*, P_4^*, P_5^*\}$.

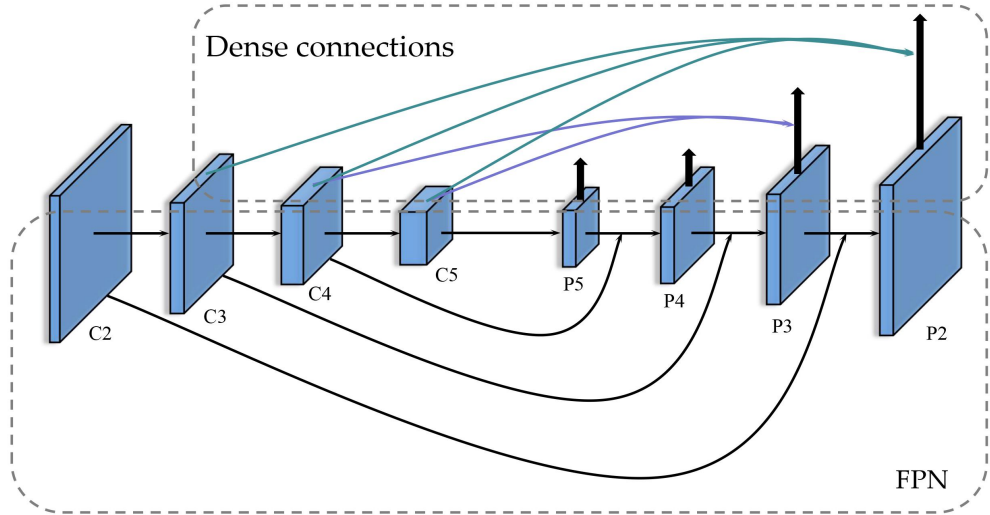

Figure 2. DMFFPN includes dense connection parts to extract more sufficient features.

Figure 3 gives an example to get $P_2^*$, we firstly get $P_2$ by following the initial FPN step, and then feature maps $\{C_3, C_4, C_5\}$ reduce its number of channel to 256 by using 1x1 convolutional layers and reshape it into the same size. Next, we merge all feature maps from $\{C_3, C_4, C_5, P_2\}$ by utilizing the operation of concatenation rather than a simple addition to getting stronger representation, the comparison results of two different fusion method of feature maps can be referred to the Section 4.1. Finally, we eliminate the aliasing effects trough a 3x3 convolutional layer that the output number of channels is 256.

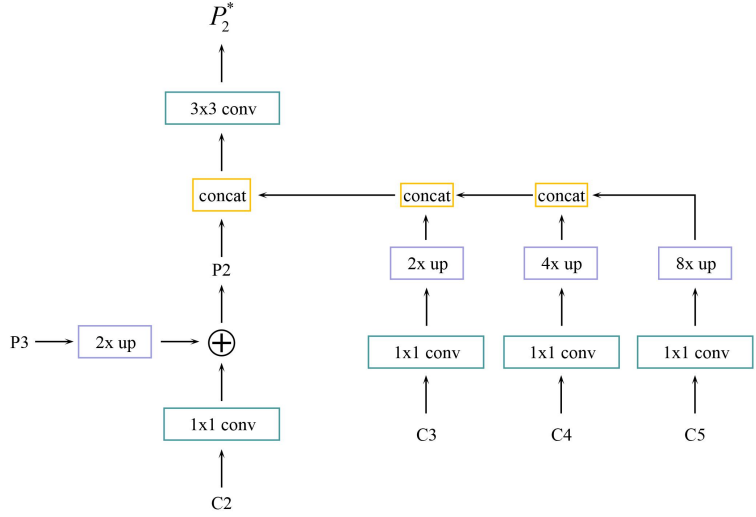

Figure 3. Detailed process of prediction

### 3.2 Loss Function

The multi-task loss is used and defined as follows.

$$L = \frac{1}{N}\sum_{t=1}^{T}\sum_{n=1}^{N} L_{cls}\left(c_{tn}, c_{tn}^{*}\right) + \frac{\lambda}{N}\sum_{t=1}^{T}\sum_{n=1}^{N} L_{reg}\left(r_{tn}, r_{tn}^{*}\right) \tag{5}$$

Here, $L_{cls}$ and $L_{reg}$ respectively represents the loss function of classification and bounding box regression at each stage $t$. Note that, softmax cross-entropy is defined as the classification loss, bounding box regression loss is defined as smooth L1 loss which follows the setting in Faster R-CNN. T is the total number of the cascaded stage. $\lambda$ controls the balance between the different task, we set $T=3$, $\lambda=1$ in this paper. In addition, the function $L_{cls}$ and $L_{reg}$ are respectively defined as follow:

$$L_{cls}\left(c_{tn}, c_{tn}^{*}\right) = -c_{tn}^{*}\log c_{tn} \tag{7}$$

$$L_{reg}\left(r_{tn}, r_{tn}^{*}\right) = \text{smooth}_{L_1}\left(r_{tn} - r_{tn}^{*}\right) \tag{8}$$

$$\text{smooth}_{L_1}(x) = \begin{cases} 0.5x^2, & |x| \leq 1 \\ |x| - 0.5, & x < -1 \text{ or } x > 1 \end{cases} \tag{9}$$

## 4. Experiments and Results

## 4.1 Datasets and Evaluation Criteria

### 4.1.1 Drone-based Datasets

VisDrone, Drone-based Datasets, is collected by the AISKYEYE team, Tianjin University, China. The benchmark dataset focuses on four core problems, i.e., object detection in images, object detection in videos, single object tracking, and multi-objects tracking. In this paper, we are mainly aimed at object detection in images, which include 10,209 static images(6,471 images used for training, 548 images for validation and 3,190 images for testing), 54.2k labels and 10 common objects (car, van, bus, pedestrian, tricycle, etc.) are involved.

### 4.1.2 Evaluation Criteria

Due to VisDrone has its own evaluation method, in this paper, we compare our method with state-of-art algorithms by using the corresponding criteria. Following the criteria of MS COCO, VisDrone uses $AP_{0.5:0.95}$, $AP_{0.5}$, $AP_{0.75}$, $AR_1$, $AR_{10}$, $AR_{100}$ and $AR_{500}$ metrics to evaluate the detection results. Specifically, $AP_{0.5:0.95}$ is computed by making an average value of 10 IoU thresholds from 0.5 to 0.95 with the step size 0.05. $AP_{0.5}$ and $AP_{0.75}$ are computed in a single IoU threshold 0.5 and 0.75. Note that, The max detections per images are 500, which is different from the metrics of MS COCO. Moreover, $AR_1$, $AR_{10}$, $AR_{100}$ and $AR_{500}$ are the maximum recalls of 1, 10, 100, 500 objects per images.

## 4.2 Implementation Details

All of the results of our experiments are performed on the validation set of VisDrone2019. Firstly, Tensorflow 1.12 is used as deep learning framework, and ResNets-101 is the pre-training model to initialize the network. Besides, short side of input images is resized to 800 pixels, MomentumOptimizer is selected as the optimizer, weight decay is 0.0001 and Momentum is set to 0.9. Meanwhile, we sample 256 batch size of anchors with positive-to-negative samples 1:1 at RPN stage and set the batch size of RoIs to 512 at Fast R-CNN stage where the ratio of positive and negative samples is 1:3. We train the model on a single GPU(NVIDIA GTX2070 8G) with a learning rate of 0.00125 for the first 70k iterations, 0.000125 for the next 15k iterations. flipping image randomly is used as the data augmentation. Finally, to match the objects in aerial images, we set the base size of the prior anchor to $\{16, 32, 64, 128, 256\}$ and the anchor ratio is $\{1:2, 1:1, 2:1\}$.

## 4.3 Comparative result in each category

We firstly investigate the effect of the dense connection in each category. Cascaded R-CNN with ResNets-101 as the baseline network is constructed. The detection result is listed in Table 1 for comparison. We can see that the detection accuracy of DMFFPN is 28.01% to $AP_{0.5:0.95}$, which is improved by 0.99 points compared with the initial FPN. Furthermore, the result shows that the AP in the most categories is much higher than the initial method, especially in these with smaller size objects such as bicycle, tricycle, and so on. It demonstrates a consistent improvement in the method we proposed.

Table 1. Comparative results in each category. All categories are evaluated in $AP_{0.5:0.95}$. FPN: Feature Pyramid Networks. DMFFPN: Dense Multiscale Feature Fusion Pyramid Networks.

| Baseline | Method | $AP_{0.5:0.95}$[%] | ped. | person | bicycle | car | van | truck | tricycle | awn. | bus | motor |
|---|---|---|---|---|---|---|---|---|---|---|---|---|
| Cascade R-CNN | FPN | 27.0 | 25.36 | 17.99 | 9.92 | 54.66 | 34.45 | 26.81 | 17.45 | 11.43 | **44.2** | 23.85 |
| | DMFFPN | **28.01** | **26.02** | **18.47** | **11.71** | **55.4** | **35.76** | **27.38** | **20.51** | **11.86** | 43.59 | **24.6** |

## 4.4 Ablation Study

**Effect of dense connection.** As discussed in sec.3.1, the dense connection structure is beneficial to enhance contextual semantic information for objects, especially for small objects. In Table 2 we explore the effect of dense connection with different pyramid layer P3 and P2, the experiment shows that $AP_{0.5:0.95}$ is 27.8% when only using dense connection with P3 which is increased by 0.8 points compared with initial structure, and $AP_{0.5:0.95}$ is 27.87% when only using dense connection with P2, which is increased by 0.87 points. It indicates that these two connection methods can both more effectively integrate the low-level and high-level information and provide more advanced feature information for object detection.

**Concatenation and addition.** The experiment also makes a comparison on two different fusion method of feature maps, we can see from Table 2 that $AP_{0.5:0.95}$ of the addition operation is 27.7%, the Concatenation operation is 28.01%. It suggests that the fusion method of concatenation is more beneficial for detecting objects in aerial image, but there is also more occupation of computing resources when using such a method in networks.

Table 2. Ablation study on MPFPN. Add: the operation of addition. Concat: the operation of concatenation.

| Fusion Strategy | Dense connection with P3 | Dense connection with P2 | $AP_{0.5:0.95}$[%] | $AP_{0.5}$[%] | $AP_{0.75}$[%] |
|---|---|---|---|---|---|
| Concat | √ | × | 27.8 | 53.24 | 25.4 |
|  | × | √ | 27.87 | 53.34 | 25.32 |
|  | √ | √ | **28.01** | **53.55** | **25.6** |
| Add | √ | √ | 27.7 | 52.96 | 25.19 |

## 4.5 Comparison with the state-of-the-art

We perform a comparison with the existing popular object detection method on VisDrone validation set. The experiment result shows that our method reaches the best performance in all method for AP and AR as shown in Table 3. Note that no additional data augmentation methods or training tricks, such as multiscale training and bigger backbone, are used in the experiment.

Table 3. Comparison with the state-of-the-art method.

| Method | $AP_{0.5:0.95}$[%] | $AP_{0.5}$[%] | $AP_{0.75}$[%] | $AR_1$[%] | $AR_{10}$[%] | $AR_{100}$[%] | $AR_{500}$[%] |
|---|---|---|---|---|---|---|---|
| FPN[16] | 25.56 | 50.21 | 23.87 | 0.47 | 5.07 | 29.71 | 39.87 |
| Cascade R-CNN[23] | 25.09 | 49.93 | 24.02 | 0.48 | 4.83 | 30.38 | 43.24 |
| DetNet59[25] | 24.28 | 48.24 | 23.36 | 0.45 | 4.61 | 29.87 | 37.28 |
| RefineDet[26] | 23.93 | 46.78 | 23.08 | 0.45 | 4.45 | 27.14 | 40.58 |
| RetinaNet[19] | 20.84 | 39.37 | 19.63 | 0.41 | 3.23 | 23.35 | 34.27 |
| YOLOv3[9] | 17.15 | 36.25 | 14.4 | 0.64 | 5.08 | 21.77 | 26.04 |
| ours | 28.01 | 53.55 | 25.6 | 0.58 | 5.88 | 34.28 | 43.84 |

Figure 5 gives the visual result of the proposed framework compared with the ground-truth. It can be seen that our method can effectively process the most objects with different scale and viewpoint.

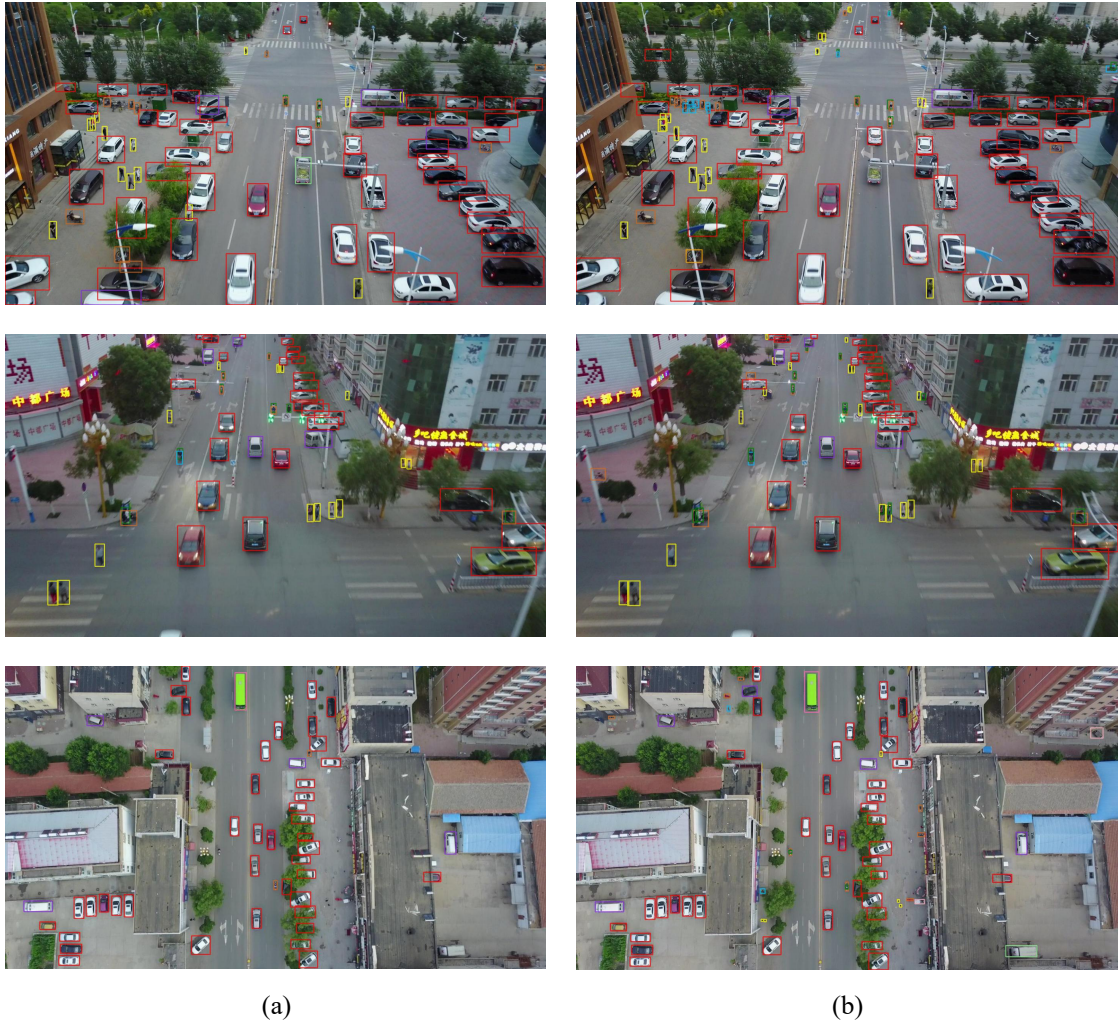

(a)                  (b)

Figure 5. Visualization result on VisDrone validation datasets. (a) detected results of the proposed framework (b) Ground-truth.

## 5. Conclusion

In this paper, we have proposed an end-to-end object detection framework, which fully utilizes the multi-scale feature information as much as possible, which is beneficial to detect small objects in aerial images. Meanwhile, the Cascade architecture in the second stage refines the bounding box regression to enhance the localization capability for objects. Experiment on VisDrone datasets demonstrates the effectiveness of the framework proposed and reaches a state-of-the-art performance in object detection for UAV-captured images. Despite achieving good performance in most classes, there are still some issues that categories like awning-tricycle, bicycle, reached a much lower result than others. One possible reason is that the number of its samples is not enough to be trained well and heavy occlusion happened in these categories. We need to explore the methods of resolution in the future.